\documentclass[10pt,twocolumn,letterpaper]{article}

\usepackage{iccv}
\usepackage{times}
\usepackage{epsfig}
\usepackage{graphicx}
\usepackage{amsmath}
\usepackage{amssymb}

\usepackage{cite}
\usepackage{makecell}

\usepackage[pagebackref=true,breaklinks=true,letterpaper=true,colorlinks,bookmarks=false]{hyperref}

\iccvfinalcopy % *** Uncomment this line for the final submission

\def\iccvPaperID{7} % *** Enter the ICCV Paper ID here

\ificcvfinal\fi

\begin{document}

%%%%%%%%% TITLE
\title{An extensive lab- and field-image dataset of crops and weeds for computer vision tasks in agriculture}

\author{Michael A. Beck\\
University of Winnipeg\\
Winnipeg, MB Canada\\
{\tt\small m.beck@uwinnipeg.ca}
\and
Chen-Yi Liu\\
University of Winnipeg\\
Winnipeg, MB Canada\\
\and
Christopher P. Bidinosti\\
University of Winnipeg\\
Winnipeg, MB Canada\\
\and
Christopher J. Henry\\
University of Winnipeg\\
Winnipeg, MB Canada\\
\and
Cara M. Godee\\
University of Winnipeg\\
Winnipeg, MB Canada\\
\and
Manisha Ajmani\\
University of Winnipeg\\
Winnipeg, MB Canada\\
}

\maketitle
\ificcvfinal\thispagestyle{empty}\fi

%%%%%%%%% ABSTRACT
\begin{abstract}
We present two large datasets of labelled plant-images that are suited
towards the training of machine learning and computer vision models.
The first dataset encompasses as the day of writing over 1.2 million
images of indoor-grown crops and weeds common to the Canadian Prairies and many US states. 
The second dataset consists of over 540,000 images of plants
imaged in farmland. All indoor plant images are labelled by species
and we provide rich metadata on the level of individual images. This comprehensive database
allows to filter the datasets under user-defined specifications such
as for example the crop-type or the age of the plant. Furthermore,
the indoor dataset contains images of plants taken from a wide variety
of angles, including profile shots, top-down shots, and angled perspectives.
The images taken from plants in fields are all from a top-down perspective
and contain usually multiple plants per image. For these images metadata
is also available. In this paper we describe both datasets' characteristics with respect
to plant variety, plant age, and number of images.
We further introduce an open-access sample of the indoor-dataset that
contains 1,000 images of each species covered in our dataset. These,
in total 14,000 images, had been selected, such that they form a representative
sample with respect to plant age and individual plants per species.
This sample serves as a quick entry point for new users to the dataset,
allowing them to explore the data on a small scale and find the parameters
of data most useful for their application without having to deal with hundreds 
of thousands of individual images.
\end{abstract}

%%%%%%%%% BODY TEXT
\section{Introduction\label{sec:Introduction}}

A sufficient amount of labelled data is critical for machine-learning
based models and a lack of training data often forms the bottleneck
in the development of new algorithms. This problem is magnified in
the area of digital agriculture as the objects of interest -- plants
-- have a wide variety in appearance that stems from the plants'
growing stage, its specific cultivar, and its health. Plants also
react in appearance to outside factors such as drought, time of day,
temperature, humidity, and sunlight available. Furthermore, the correct
classification of plants requires expert knowledge, which cannot easily
be crowdsourced. All of this frames the labelling of plant-data as
a challenge that is significantly harder compared to similar image-labelling
tasks. Yet, as we
witness the introduction of sensors \cite{s16050618,8062821,ANTONACCI201895,KHANNA2019218},
robotics \cite{oberti2016advances,BECHAR2017110,BECHAR2017110part2,DBLP:journals/corr/abs-1806-06762,20193426318,RELFECKSTEIN2019100307},
and machine learning \cite{LOBET2017559,10.1371/journal.pcbi.1005993,s18082674,patricio2018computer,kamilaris2018deep,JHA20191} to
agricultural applications, there is a strong demand for such training data. This research area running under the names
of precision agriculture, digital agriculture, smart farming, or Agriculture
4.0 has the potential to increase yields while reducing hte usage of resources, such as water, fertilizer, and herbicides \cite{JHA20191,BINCH2017123,rs10111690,8387426,barbedo2013digital,FAHLGREN201593,SINGH2016110,SHAKOOR2017184,https://doi.org/10.3732/ajb.1700044,giuffrida2018citizen,TARDIEU2017R770,BACCO2019100009,CHARANIA2020100142}.
This next revolution in agriculture is fuelled by data, in particular
labelled image-data with rich metadata information. 

\begin{table}
\begin{centering}
\caption{Overview on available datasets.\label{tab:overview}}
\par\end{centering}
\centering{}%
\begin{tabular}{|c|c|c|}
\hline 
Dataset & Type & Image Count\tabularnewline
\hline 
\hline 
\emph{Lab-data} & \makecell{Indoor grown \\ crops and weeds} & \makecell{1.2 million images}\tabularnewline
\hline 
Subsample & \makecell{Sample \\ of \emph{Lab data}} & \makecell{14,000 images}\tabularnewline
\hline
\hline 
\emph{Field-data} & \makecell{Crops and weeds \\ in farmland} & \makecell{540,000 images}\tabularnewline
\hline
\end{tabular}
\end{table}

In this paper we describe two datasets, each consisting of hundreds
of thousands of images, suitable for machine-learning and computer
vision applications. The first dataset, the \emph{lab-data}, consists
of indoor-grown crops and weeds that had been imaged from a wide variety
of angles. The plants selected are species common on farmlands in the Canadian prairies and many US states. 
This dataset consists of images of individual plants, images
showing several plants with and without bounding boxes, and metadata
for each image and plant. At the time of writing more than 1.2 million
images have been added to this dataset, since April 2020.
All of these images had been captured and automatically labelled
by using a robotic system as in \cite{10.1371/journal.pone.0243923}. The
second dataset, the \emph{field-data}, consists of images taken in
the field in the growing seasons of 2019 and 2020. These images show
a top down perspective on the crops (and weeds) as they grow in cultivated
farmland. 

We present here these two datasets with respect to some
of their key metrics, such as the number of images per species imaged.
Further we provide a sample from the lab-data consisting of
14,000 images. This sample (around 1\% of the total data) is intended
to give researchers an overview on how the data is structured with
respect to available plant ages and growing stages, as well as the
number of individual plants imaged by the system. For this, we carefully
selected images from the entire dataset, such that the age distribution
per species is preserved and a wide range of individual plants is
represented. Following best-practices for data-accessability, as outlined
in \cite{LOBET2017559}, we give immediate open access to this data-sample, 
provide the metadata, and give insights on how the data was collected.
In terms of long-term data storage and accessability both full datasets will eventually 
be fully open-source following a data management plant of stagewise release. 
Our goal is to provide researchers in digital agriculture with labeled data to facilitate
data-driven innovation in the field.

The rest of this paper is structured as follows: In Chaper \ref{sec:Structure}
we describe the general structure of both datasets and their metadata.
In Chapter \ref{sec:Characteristics} we visualize and describe key
characteristics of the datasets, such as the number of individual
plants per species. In Chapter \ref{sec:subsamples} we describe the
structure of the sample. We conclude the paper in Chapter \ref{sec:Conclusion}
with information on the planned growth of the datasets and how to obtain the sample and the original dataset.

%------------------------------------------------------------------------
\section{Data- and Metadata-Structure \label{sec:Structure}}

\begin{figure}
\begin{centering}
\includegraphics[width=0.9\columnwidth]{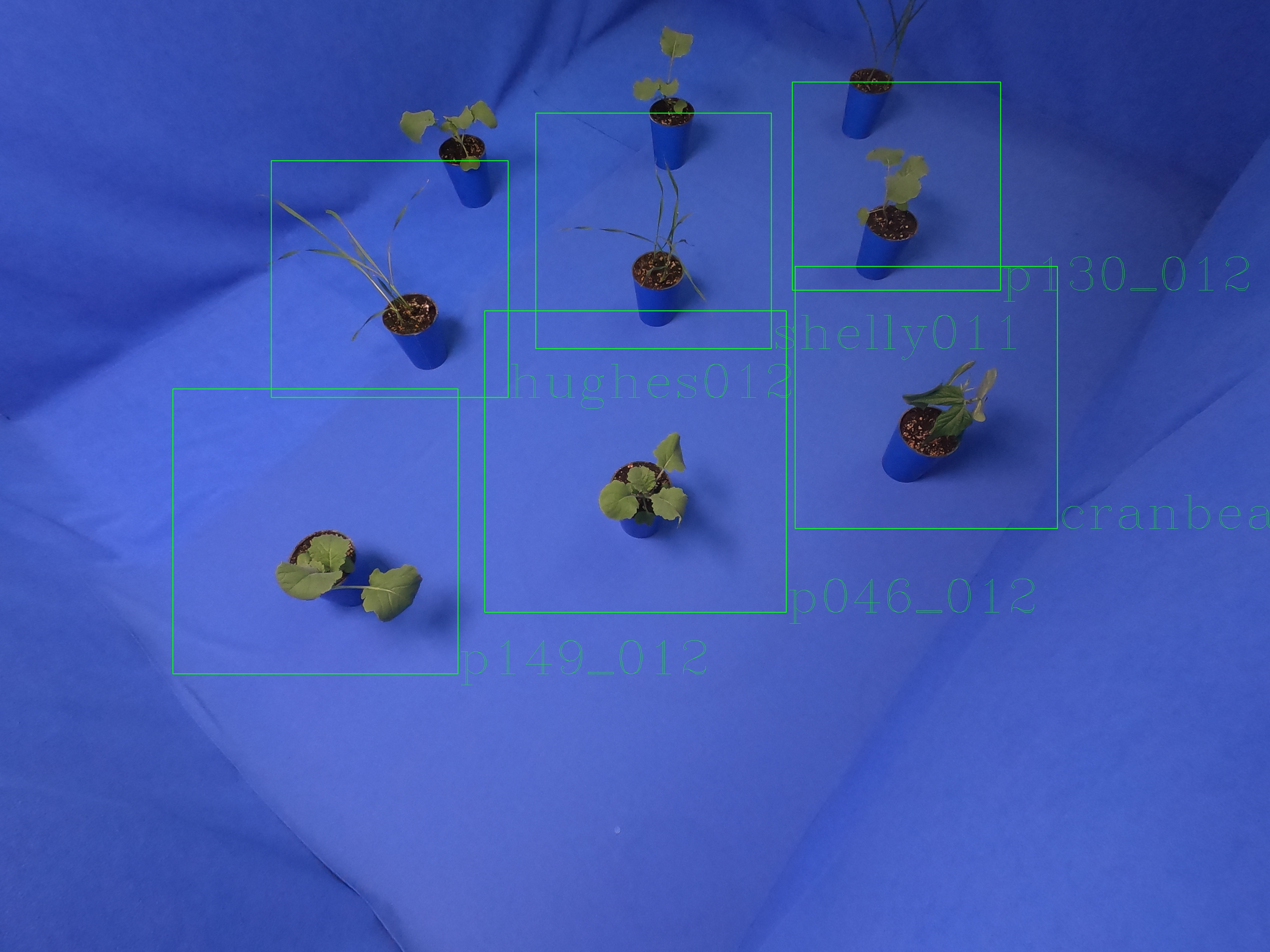}\smallskip{}
\par\end{centering}
%\begin{centering}
%\includegraphics[width=0.9\columnwidth]{images/20200626120042-pose10-bb}\smallskip{}
%\par\end{centering}
\begin{centering}
\includegraphics[width=0.45\columnwidth]{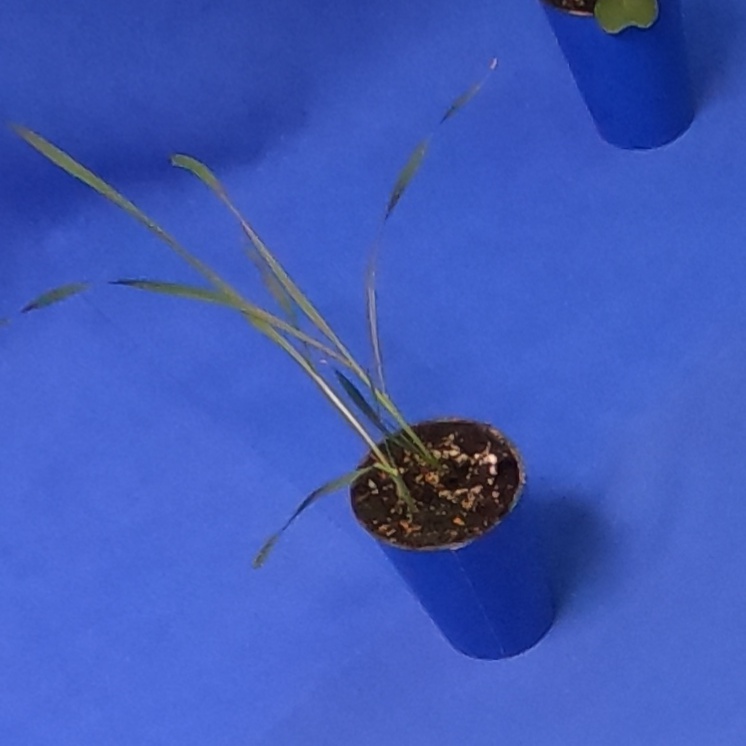}\includegraphics[width=0.45\columnwidth]{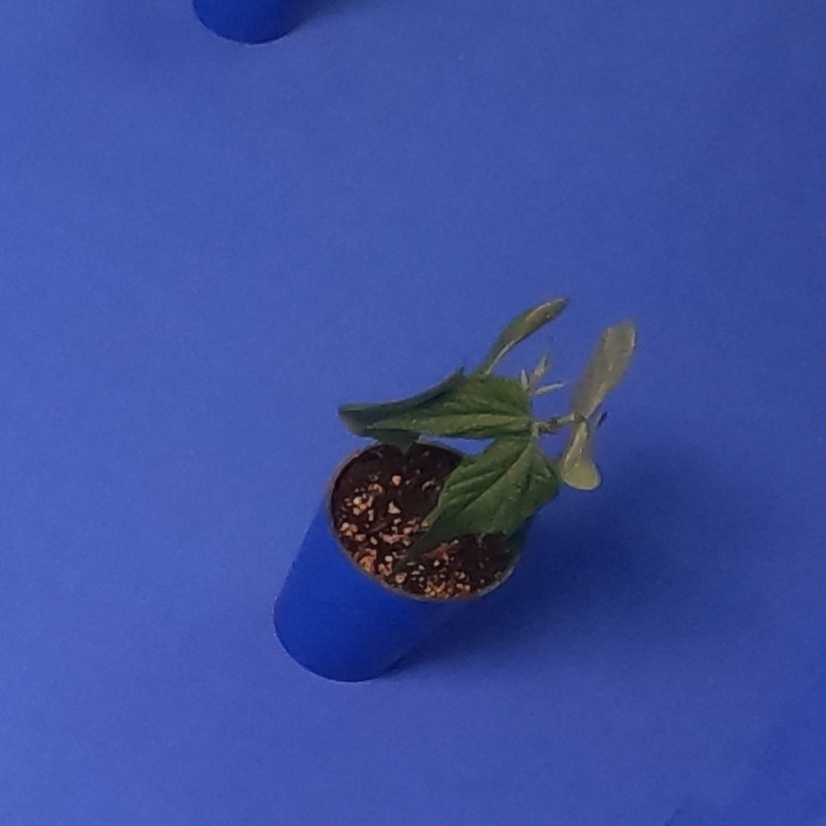}\smallskip{}
\par\end{centering}
\begin{centering}
\includegraphics[width=0.45\columnwidth]{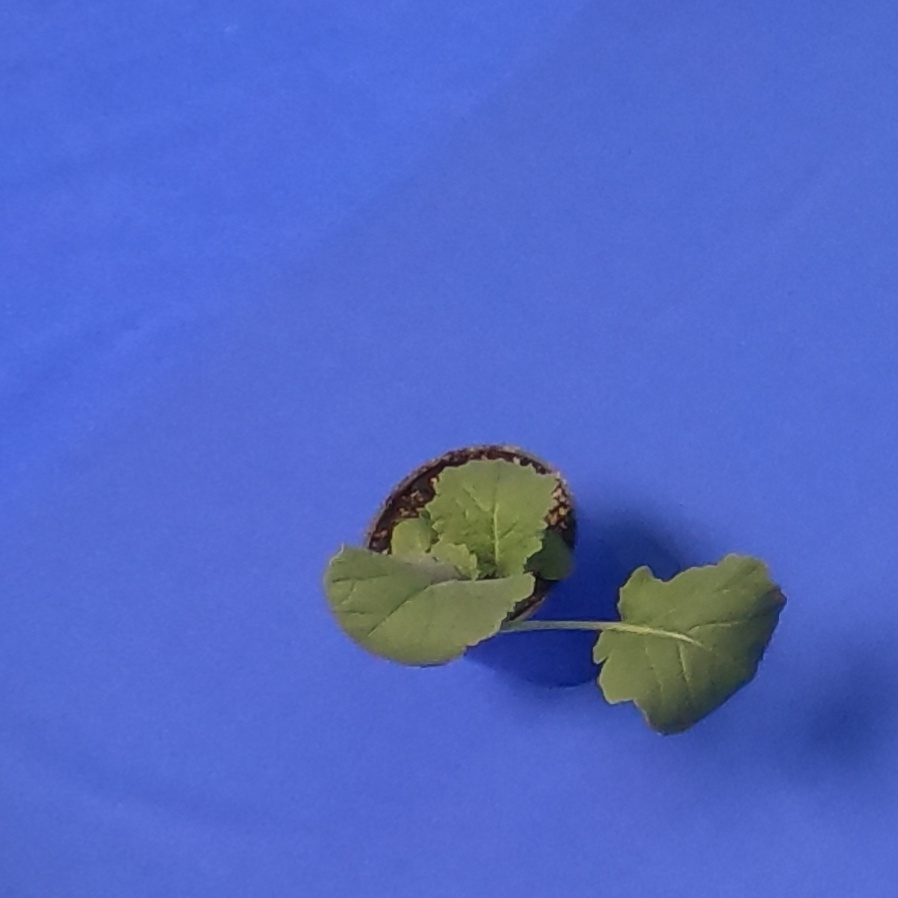}\includegraphics[width=0.45\columnwidth]{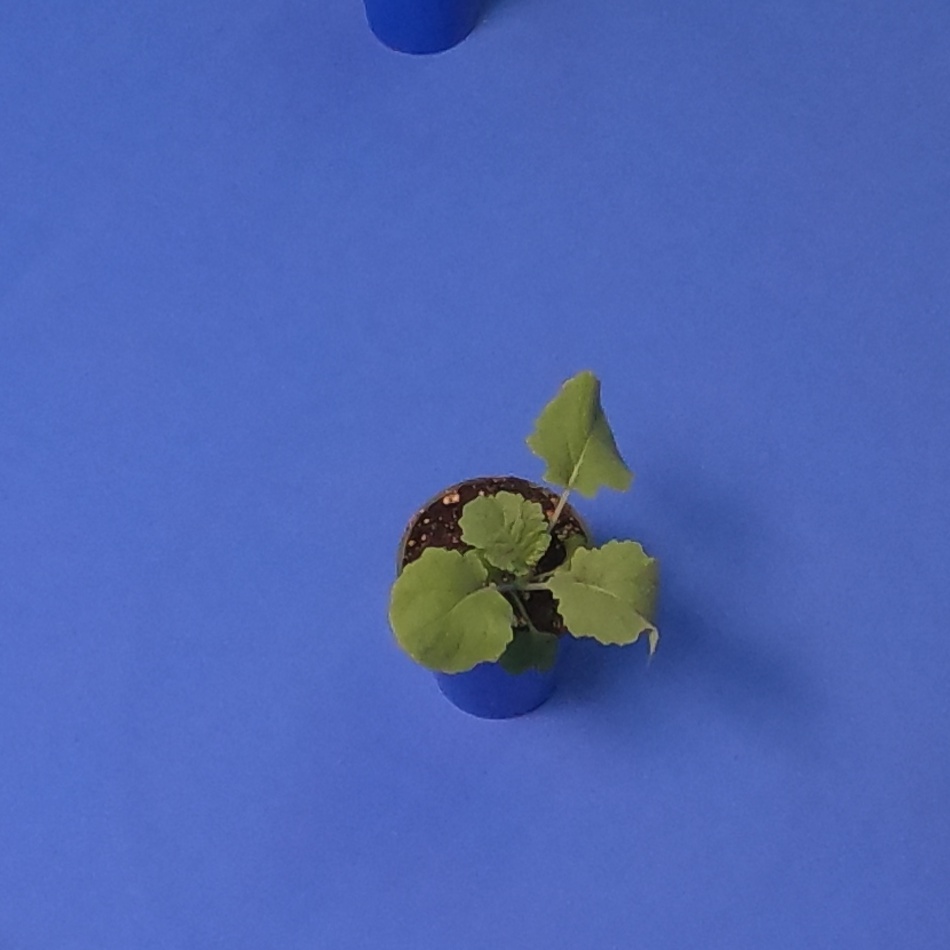}
\par\end{centering}
\caption{Two images taken by the system with drawn calculated bounding boxes.
Below four single plant images cropped from the first image.\label{fig:Example}}

\end{figure}

The lab-data can be divided into 4 different kind of files that relate
to each other as follows:
\begin{itemize}
\item Plain images: These are the images as captured by the camera. They
typically show several plants in the same image.
\item Bounding box images: These images are the same as the original images
with the difference that they are overlaid with visible bounding boxes
around the plants as calculated by the system. Plants too close to
the border of the image or overlapping too far into each other are
not being bounded by the system.
\item Single plant images: These are images cropped out from plain images
according to the calculated bounding boxes. Only plants for which
a bounding box has been drawn are cropped out as individual
image.
\item JSON-files: These files contain the metadata associated with each
plain image and are described in more detail below.
\end{itemize}
See Figure \ref{fig:Example} for examples on a plain image, the respective
image with bounding boxes, and cropped out single plant images. 
\begin{figure}
\begin{centering}
\includegraphics[width=0.9\columnwidth]{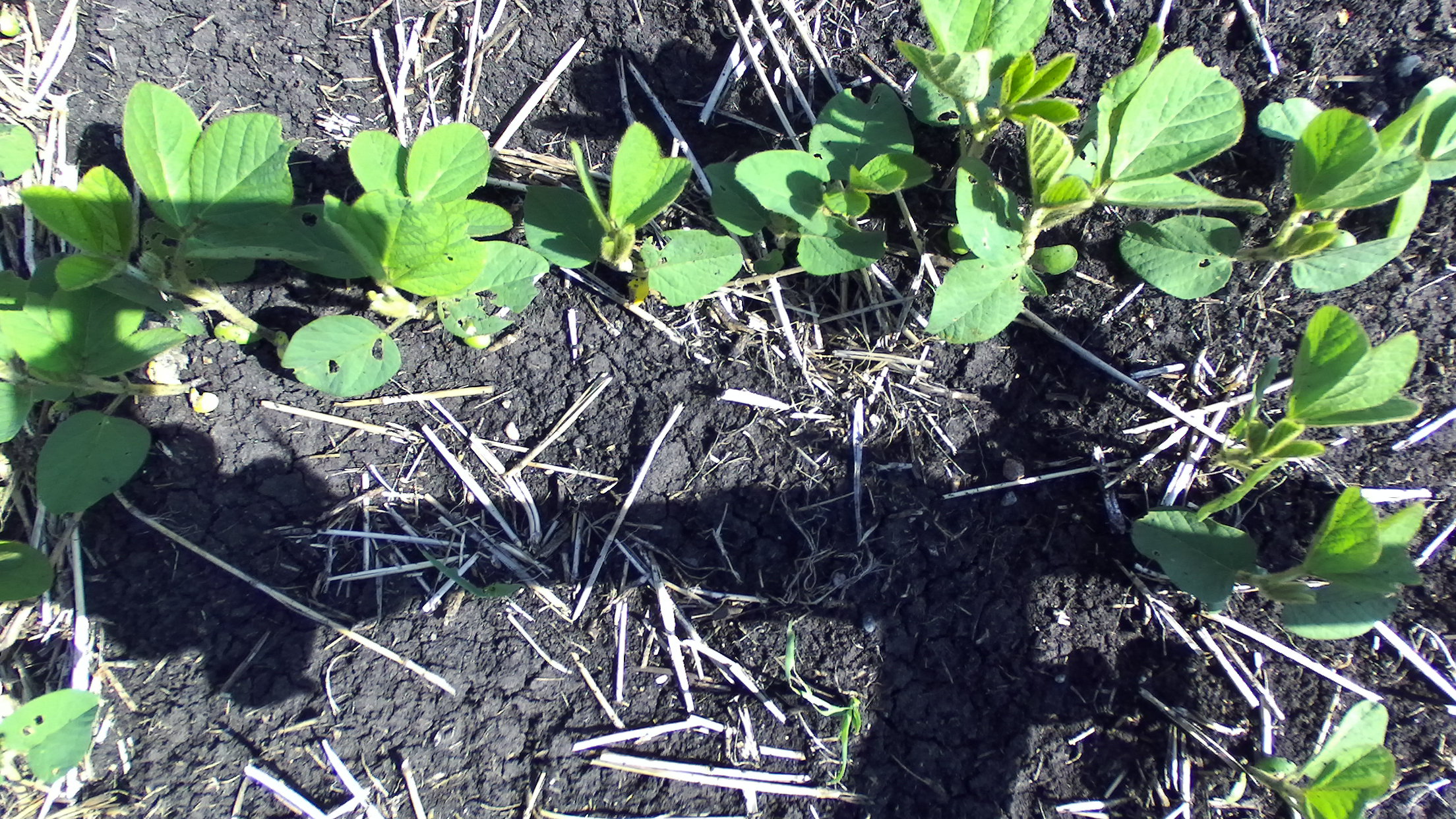}
\par\end{centering}
\caption{Example of an image taken in the field, showing several Soybean plants.\label{fig:Example-Field}}
\end{figure}

\begin{figure}

\noindent \centering{}\includegraphics[width=0.95\columnwidth]{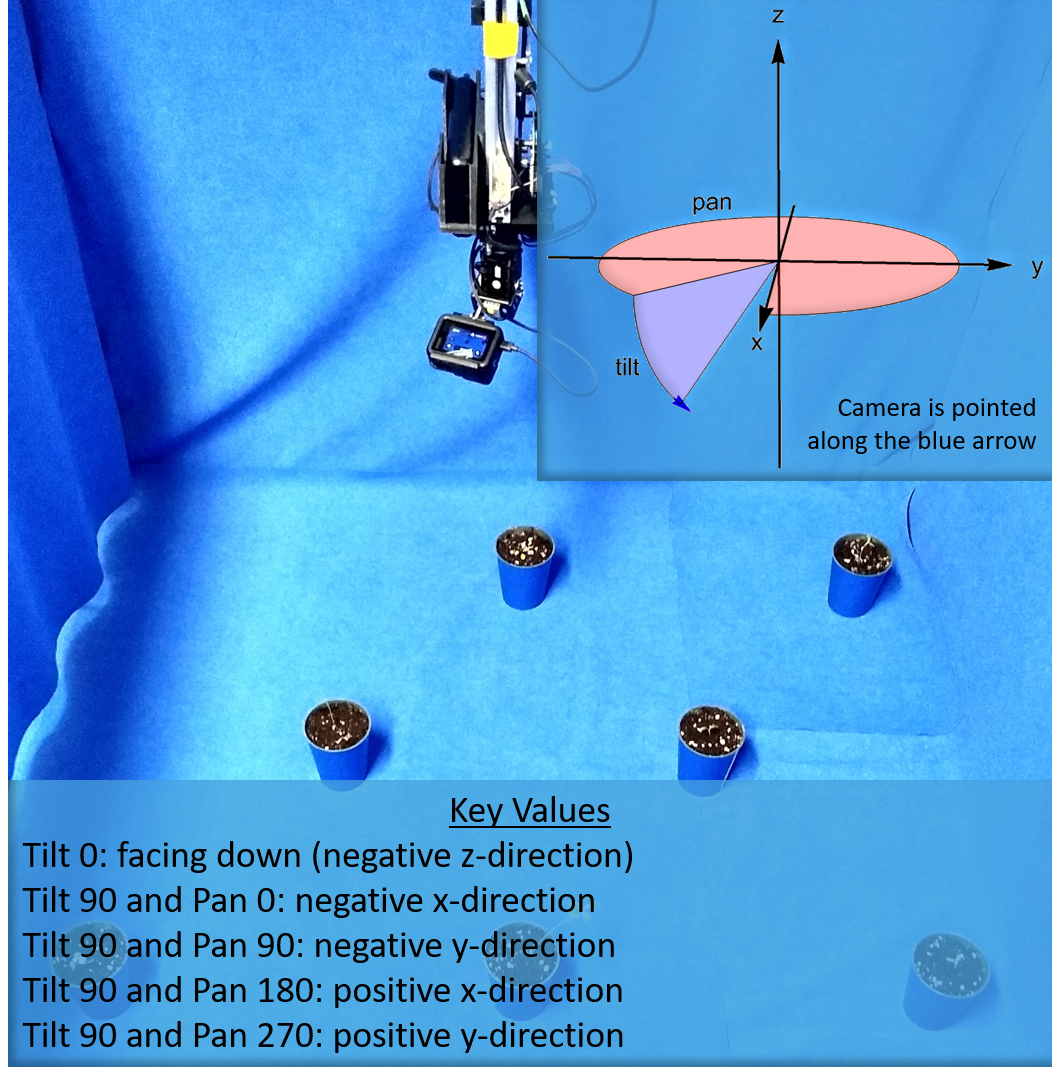}\caption{Interpretation of the coordinates x, y, z, pan, and tilt.\label{fig:coordinates}}
\end{figure}

Each JSON-file contains information about the plain image and the
respective single plant images as follows: 
\begin{itemize}
\item \emph{version: }An internally used version number 
\item \emph{file\_name: }The filename of the original image (the plain
image)
\item \emph{bb\_file\_name: }The filename of the bounding box image
\item \emph{singles, bb\_images, source\_file\_path: }These fields are internally
used filepath information for retrieving data from an object storage.
\item \emph{date: }A string containing the day of imaging in the format
YYYY-(M)M-(D)D. The string does not contain leading zeroes for months
or days. 
\item \emph{time: }A string containing the time of imaging in the format
(h)h:(m)m:(s)s. The string does not contain leading zeroes for hours,
minutes, or seconds. The timezone is CST (Central Standard Time; UTC-06:00)
and CDT (Central Daylight Time; UTC-05:00), respectively. 
\item \emph{room, institute, camera, lens:} These fields encode the location
and camera-equipment used. 
\item \emph{vertical\_res, horizontal\_res: }The resolution given in pixels
along the vertical and horizontal image-axis, respectively. 
\item \emph{camera\_pose: }This field contains the following subfields:
\emph{x, y, z, polar\_angle, azimuthal\_angle}. The first three subfields
describe the camera position in cm with respect to an origin-point
inside the imagable volume of the system. The latter two subfields
describe the camera's pan (polar\_angle) and tilt (azimuthal\_angle).
See Figure \ref{fig:coordinates} for details.
\item \emph{bounding\_boxes:} This field contains a list of elements each
cooresponding to one plant around which a bounding box was drawn by
the system (see above for the description of bounding box images and
single plant images). Each element in this list contains the same
list of subfields as follows: 
\begin{itemize}
\item \emph{plant\_id: }Each plant is identified through an individual identifier.
This allows to differentiate individual plants from the same breed
and species. 
\item \emph{label: }A common name label attached to the plant, such as ``Canola''
\item \emph{scientific\_name: }The scientific name of the plant, such as
\emph{Brassica napus}
\item \emph{subimage\_file\_name: }The filename of the respective single
plant image. This field is used to associate the single plant images
with the plain image from which they had been cropped from or to
the respective JSON-file (by replacing the file-ending from .jpg to
.json)
\item \emph{date\_planted: }The day at which the plant was planted. This information
is used to determine a plant's age, which in our case is defined
as the number of days that have elapsed between planting and imaging
\item \emph{position\_id: }An internally used identifier for the spatial
position on which the plant was located when imaged
\item \emph{x\_min, y\_min, x\_max, y\_max: }The coordinates of the upper-left
and lower-right corner of the calculated bounding box, respectively. 
\end{itemize}
\end{itemize}

The field-data collection is in structure similar to the above. However,
as there are no labels attached to individual plants we have only
the following fields in use: \emph{version, file\_name, date, time,
room, institute, camera, lens, label, vertical\_res, horizontal\_res,
source\_file\_path}. Note that the \emph{label} field is with respect
to the entire image, whereas for lab-data it was associated with a
cropped out single plant image. The entry under \emph{label} thus 
describes the type of crop that is cultivated on the farmland from which 
the image had been taken. Indeed, as imaging also took place before 
the application of herbicides, we can see weeds between the dominant 
crops on some images. Other fields in the JSON-files have the same
interpretation and usage as above. An example for a field-data image
is given in Figure \ref{fig:Example-Field}.

%------------------------------------------------------------------------
\section{Dataset Characteristics\label{sec:Characteristics}}

We now describe the lab-data with respect to different
metrics, followed by metrics on the field-data. 
\begin{figure}
\begin{centering}
\includegraphics[width=0.95\columnwidth]{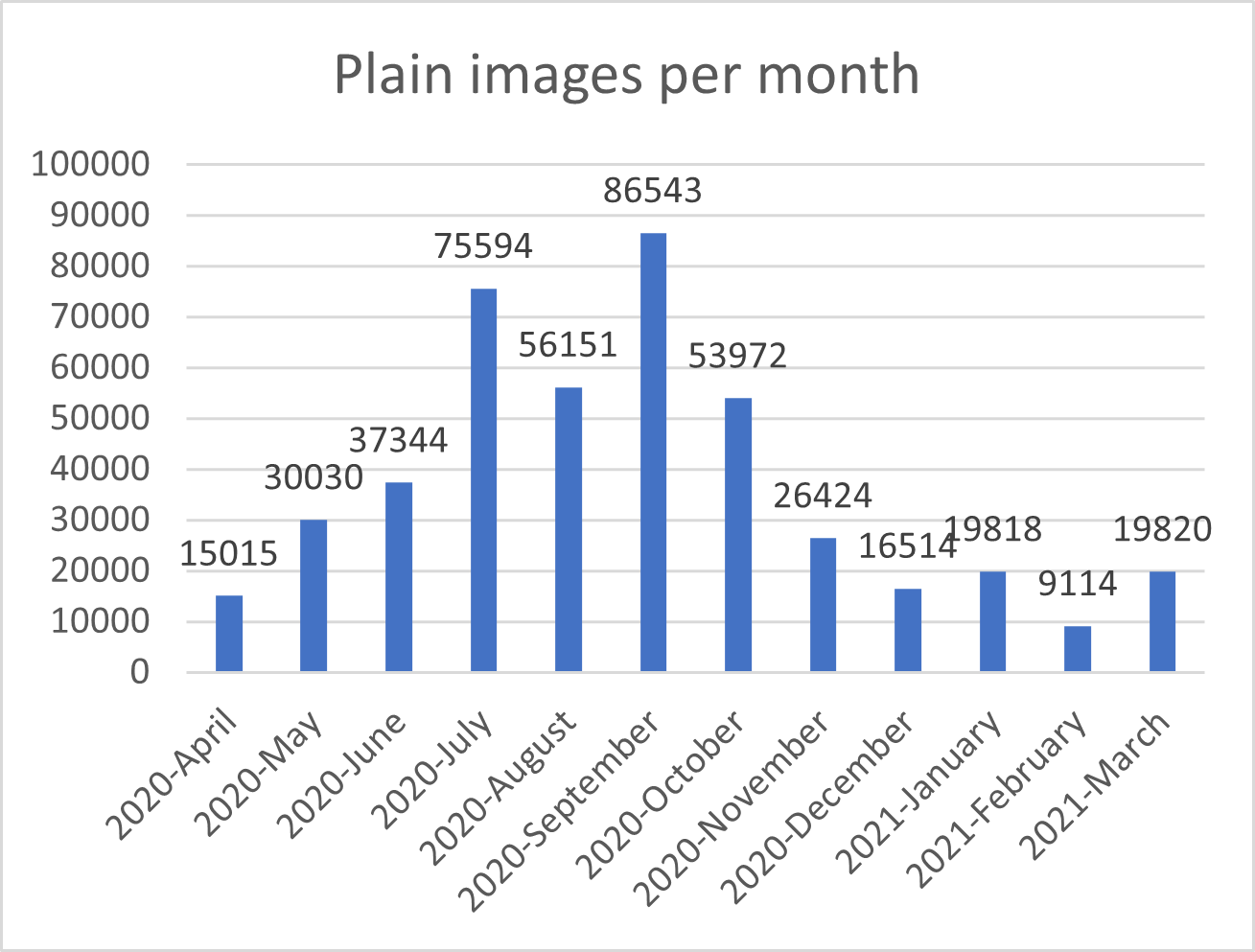}
\par\end{centering}
\begin{centering}
\includegraphics[width=0.95\columnwidth]{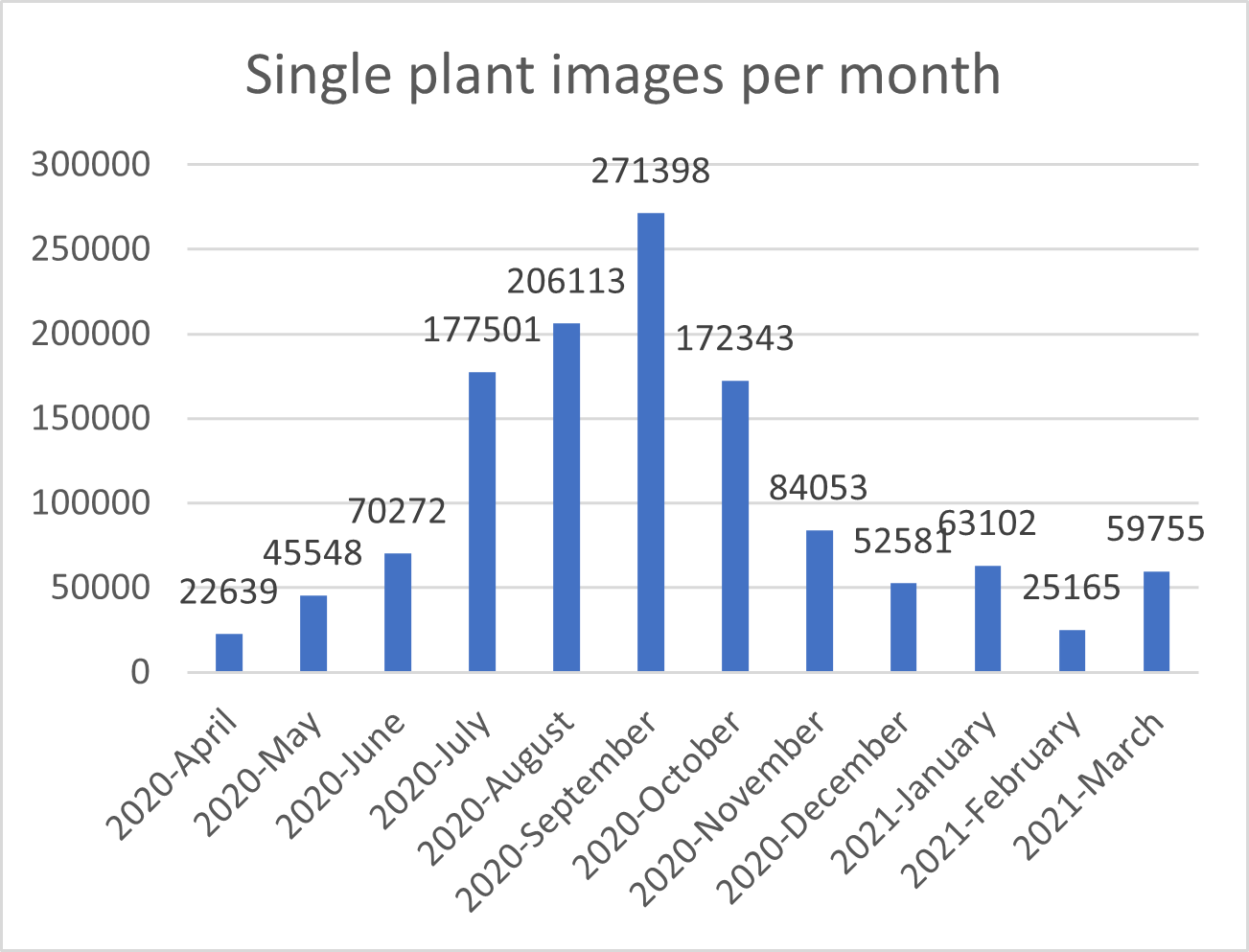}
\par\end{centering}
\begin{centering}
\includegraphics[width=0.95\columnwidth]{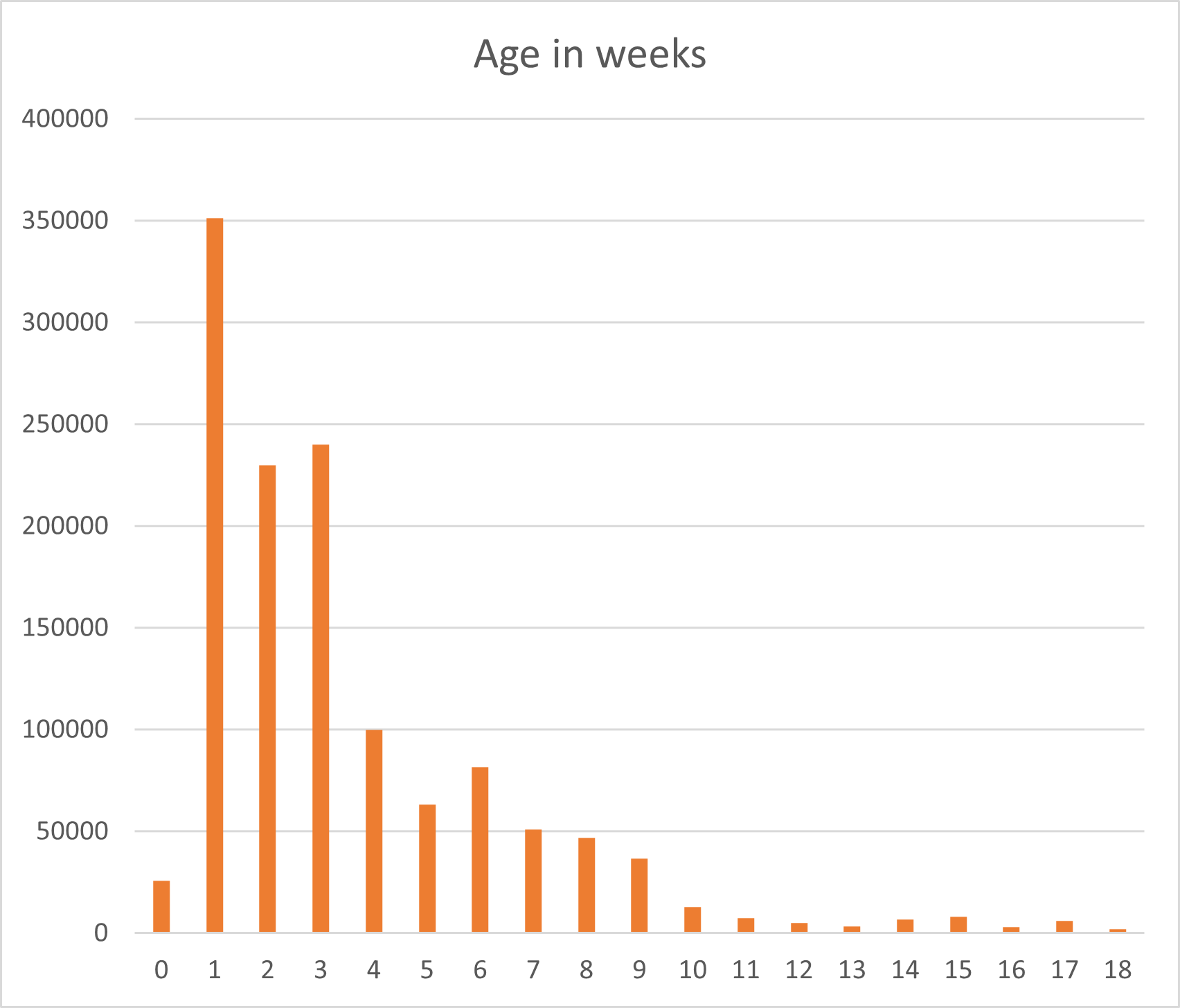}
\par\end{centering}
\caption{A: The number of plain images per month. These are images, as described
in Section \ref{sec:Structure} that contain multiple plants within
them. B: The number of single plant images per month cropped out of
the plain images. C: The age distribution of individually imaged plants,
the binning size in this histogram is 1 week.\label{fig:A:Per_month_charts} }

\end{figure}

Figure \ref{fig:A:Per_month_charts} shows the number of images taken
by the system starting from April 2020 (imaging before April 2020 took 
place and is available on request. It is not included in this dataset due to 
experimentation with the imaging system itself). As of this writing we have
taken more than 446,000 plain images (that is images that contain
multiple plants) and cropped out over 1.2 million single plant images
from these. The system is run several times per week producing thousands
of new plain images. The data acquisition rate drops significantly after 
October 2020 as access was restricted due to the COVID pandemic. We 
anticipate the acquisition rate to raise as accessability to our
facilities improve. 

The third panel in \ref{fig:A:Per_month_charts} shows the imaged
plants' age-distribution. We define age as the number of days elapsed
between seeding the individual plant and the time the image had been
taken. The histogram uses a binning size of 7 days. The majority of
plants had been imaged, when they were between 7 and 35 days old.
This corresponds to the growth stage in farmlands at which it is critical to
distinguish between weeds and crops. Thus, our emphasis on plants of 
this age matches the data needed for important applications in digital agriculture 
such as estimating germination rate of crops, quantifying weeds, and 
automated weeding.
%The reason for focusing on this early stage of the plants is due to
%more frequent overlapping of plants in later growth stages and difficulties
%to calculate bounding boxes around more mature plants. 
By our definition of
age the germination time itself influences the age-value in our metadata.
Since germination times vary between species, so do their age-distributions.
For example, the age distribution of weeds is generally shifted towards
``older'' plants by one or more weeks compared to crops. Indeed,
in our efforts to grow both of them, we have encountered that weeds
require a longer time and more care to germinate. Most plants are only
imaged towards a point at which they ``outgrow'' the system, i.e.,
where the plants' size and shape lead to overlaps and inaccuracies
when calculating bounding boxes.

Table \ref{tab:SP_per_species} lists how many single plant images
per species are in the dataset. The number varies strongly by species,
which is due to availability of seeds, germination success, and access 
to our facilities. 

In Table \ref{tab:ind_per_species} we list how many individual plants
we have imaged per species. Again the numbers vary due to availability and 
germination success of seeds and access to our facilities. 

\begin{table}
\begin{centering}
\caption{Number of single plant images per species.\label{tab:SP_per_species}}
\par\end{centering}
\begin{centering}
\begin{tabular}{|c|c|c|}
\hline 
Common Name & Scientific Name & \thead{Image \\ count}\tabularnewline
\hline 
\hline 
Barley & \emph{Hordeum vulgare} & 30597\tabularnewline
\hline 
Barnyard Grass & \emph{Echinochloa crus-galli} & 76258\tabularnewline
\hline 
Common Bean & \emph{Phaseolus vulgaris} & 159217\tabularnewline
\hline 
Canada Thistle & \emph{Cirsium arvense} & 89731\tabularnewline
\hline 
Canola & \emph{Brassica napus} & 255004\tabularnewline
\hline 
Dandelion & \emph{Taraxacum officinale} & 87426\tabularnewline
\hline 
Field Pea & \emph{Pisum sativum} & 68658\tabularnewline
\hline 
Oat & \emph{Avena sativa} & 59153\tabularnewline
\hline 
Smartweed & \emph{Persicaria spp.} & 99650\tabularnewline
\hline 
Soybean & \emph{Glycine max} & 203980\tabularnewline
\hline 
Wheat & \emph{Triticum aestivum} & 120417\tabularnewline
\hline 
Wild Buckwheat & \emph{Fallopia convolvulus} & 24973\tabularnewline
\hline 
Wild Oat & \emph{Avena fatua} & 7065\tabularnewline
\hline 
Yellow Foxtail & \emph{Setaria pumila} & 14815\tabularnewline
\hline 
\end{tabular}
\par\end{centering}
\end{table}

\begin{table}
\begin{centering}
\caption{Number of individual plants per species.\label{tab:ind_per_species}}
\par\end{centering}
\centering{}%
\begin{tabular}{|c|c|c|}
\hline 
Common Name & Scientific Name & \thead{Individual \\ plants}\tabularnewline
\hline 
\hline 
Barley & \emph{Hordeum vulgare} & 14\tabularnewline
\hline 
Barnyard Grass & \emph{Echinochloa crus-galli} & 21\tabularnewline
\hline 
Common Bean & \emph{Phaseolus vulgaris} & 53\tabularnewline
\hline 
Canada Thistle & \emph{Cirsium arvense} & 14\tabularnewline
\hline 
Canola & \emph{Brassica napus} & 128\tabularnewline
\hline 
Dandelion & \emph{Taraxacum officinale} & 16\tabularnewline
\hline 
Field Pea & \emph{Pisum sativum} & 24\tabularnewline
\hline 
Oat & \emph{Avena sativa} & 28\tabularnewline
\hline 
Smartweed & \emph{Persicaria spp.} & 21\tabularnewline
\hline 
Soybean & \emph{Glycine max} & 84\tabularnewline
\hline 
Wheat & \emph{Triticum aestivum} & 47\tabularnewline
\hline 
Wild Buckwheat & \emph{Fallopia convolvulus} & 15\tabularnewline
\hline 
Wild Oat & \emph{Avena fatua} & 3\tabularnewline
\hline 
Yellow Foxtail & \emph{Setaria pumila} & 5\tabularnewline
\hline 
\end{tabular}
\end{table}

We now give a short description of the field-data. The collection
of field-data was performed by imaging the field via a stereoscopic
camera mounted to a tractor. The camera, pointed straight down, records
a video as the tractor drives through the field. We chose one of the
two video channels to extract frames as images. These images form the field-data
collection. Here the amount of images extracted is chosen such that
consecutive images show some overlap with respect ot the area imaged. 
We also provide the video data itself for the user to extract images under 
their own timing conditions
or to work on the video directly. Table \ref{tab:field_by_month} and
Table \ref{tab:field_by_crop} give a breakdown by the number of images
extracted from the videos per month and species, respectively. Field-data from 
the 2020 growing season is further accompanied by metadata information about 
temperature, wind-speed, cloud coverage, and camera height from ground.

\begin{table}
\begin{centering}
\caption{Number of field images per month.\label{tab:field_by_month}}
\par\end{centering}
\centering{}%
\begin{tabular}{|c|c|c|}
\hline 
Year & Month & Image count\tabularnewline
\hline 
\hline 
2019 & June & 45954\tabularnewline
\hline 
 & July & 84033\tabularnewline
\hline 
\hline 
2020 & May & 14084\tabularnewline
\hline 
 & June & 197980\tabularnewline
\hline 
 & July & 167896\tabularnewline
\hline 
 & August & 32230\tabularnewline
\hline 
\end{tabular}
\end{table}

\begin{table}
\begin{centering}
\caption{Number of field images per crop.\label{tab:field_by_crop}}
\par\end{centering}
\centering{}%
\begin{tabular}{|c|c|c|}
\hline 
Common Name & Scientific Name & \thead{Individual \\plants}\tabularnewline
\hline 
\hline 
Canola & \emph{Brassica napus} & 137551\tabularnewline
\hline 
Faba bean & \emph{Vicia faba} & 24147\tabularnewline
\hline 
Oat & \emph{Avena sativa} & 18578\tabularnewline
\hline 
Soybean & \emph{Glycine max} & 342780\tabularnewline
\hline 
Wheat & \emph{Triticum aestivum} & 19121\tabularnewline
\hline 
\end{tabular}
\end{table}

\section{Description of subsample\label{sec:subsamples}}

To create a visual overview on the lab-data we created a subsample
of it that is structured as follows:

For each species listed in Table \ref{tab:SP_per_species} we have
selected 1,000 single plant images, thus the subsample contains 14,000
images. Furthermore, within each of these categories we have selected
images, such that the age-distribution of the 1,000 images closely
matches the age-distribution of all available images for that species.
In addition we selected images such that all individual plants grown
are represented in the subsample with the following exceptions: There
are 51 individual Common Beans present in the subsample (instead of
53 in the entire dataset), as well as 113 Canola plants (of 128),
51 Soybean plants (of 84) and 37 Wheat plants (of 47). The distribution
of the image dimensions (width, height) for the subsamples resembles
the size distribution of the entire dataset, we did however not select
images to directly optimize the sample under that criteria. 

The total size on disk of the subsample is approximately 2.2 GB. The subsample
does only contain single plant images, which are organized in one
subfolder per species. We consider this subsample as a good entry
point into the entirety of the dataset, which can be used 
to train some initial models. For example, simple models that differentiate between  
species or classes of species (e.g., monocots versus dicots,
crops versus weeds). 

\section{Conclusion and data availability\label{sec:Conclusion}}

In this paper we presented an extensive dataset of labelled plant
images. These images show crops and weeds as common in the Canadian prairies and northern US states. 
After describing the data-structure we presented a subsample
that mirrors the full dataset in key characteristics, but is smaller in overall size and thus more tractable. We are actively growing
the dataset into several dimensions: New field- and lab-data is being
acquired and processed as of writing. Furthermore, additional
data-sources such as the generation of 3d-pointclouds and hyperspectral scans are being
tested and developed. Additional field-data sources are also being
explored, including imagery from UAVs and a semi-autonomous rover.
Data from these sources will accompany the datasets presented in this
paper in the near future. 

The 14,000 images sample is available on  \href{https://doi.org/10.25739/rwcw-ex45}{https://doi.org/10.25739/rwcw-ex45}
at the CyVerse Data Store, a portal for full data lifecycle management. The full
dataset which contains 1.2 million single plant images (and counting)
is made available to researchers and industry through the data-portal hosted by EMILI under 
\href{http://emilicanada.com/}{http://emilicanada.com/} (Digital Agriculture Asset Map). The authors take Lobet's general critique 
\cite{LOBET2017559} on data-driven research in digital agriculture (or any research field) seriously. We further created a datasheet following the guidelines of Gebru et al. \cite{gebru2020datasheets}

{\small
\bibliographystyle{ieeetr}
\bibliography{Refs/references}
}

\end{document}